\pdfoutput=1

\documentclass{ifacconf}

\usepackage[export]{adjustbox}

\usepackage{graphics}      
\usepackage{natbib}        

\usepackage[dvipsnames]{xcolor}
\definecolor{myorange}{RGB}{245,156,74}
\usepackage{soul}

\usepackage{amsmath}
\usepackage{mathtools}
\usepackage{amsfonts}
\usepackage{amssymb}

\usepackage{tabularx}

\usepackage{array}

\usepackage{multirow, multicol}
\usepackage{siunitx}

\makeatletter
\def\endfigure{\end@float}
\def\endtable{\end@float}
\makeatother

\let\ifacconfcaptionwidth\captionwidth
\usepackage[caption=false]{subfig}
\let\captionwidth\ifacconfcaptionwidth

\allowdisplaybreaks

    

\renewcommand{\arraystretch}{1.4}

\makeatletter
\let\old@ssect\@ssect 
\makeatother

\usepackage{natbib}
\usepackage{hyperref}
\hypersetup{
    colorlinks=true,
    linkcolor=blue,
    filecolor=magenta,
    urlcolor=blue,
    citecolor=-myorange,
}
\makeatletter
\def\@ssect#1#2#3#4#5#6{%
  \NR@gettitle{#6}
  \old@ssect{#1}{#2}{#3}{#4}{#5}{#6}
}
\makeatother

\begin{document}

\onecolumn
\begin{center}
    This paper has been accepted for publication in the \textit{11th IFAC Symposium on Intelligent Autonomous Vehicles (IAV)}.
\end{center}

\bigskip

\begin{center}
    DOI: \href{https://www.sciencedirect.com/science/article/pii/S2405896322010059?via%3Dihub}{10.1016/j.ifacol.2022.07.594}
\end{center}

\bigskip

© 2022 the authors. This work has been accepted to IFAC for publication under a Creative Commons Licence CC-BY-NC-ND.

\clearpage

\begin{frontmatter}

\title{Enough is Enough: Towards Autonomous Uncertainty-driven Stopping Criteria
\thanksref{footnoteinfo}}

\thanks[footnoteinfo]{This work was supported by the Spanish government under grant PID2019‐108398GB‐I00 and by Arag\'on government under grant DGA\_FSE T45\_20R.}

\author[First]{Julio A. Placed}
\author[First]{Jos\'e A. Castellanos}

\address[First]{\textit{Departamento de Inform\'atica e Ingenier\'ia de Sistemas, Instituto de Investigaci\'on en Ingenier\'ia de Arag\'on, Universidad de Zaragoza}, C/ Maria de Luna 1, 50018, Zaragoza, Spain.  (e-mail: \{jplaced, jacaste\}@unizar.es)}

\begin{abstract}                
Autonomous robotic exploration has long attracted the attention of the robotics community and is a topic of high relevance. Deploying such systems in the real world, however, is still far from being a reality. In part, it can be attributed to the fact that most research is directed towards improving existing algorithms and testing novel formulations in simulation environments rather than addressing practical issues of real-world scenarios. This is the case of the fundamental problem of autonomously deciding when exploration has to be terminated or changed (\textit{stopping criteria}), which has not received any attention recently. In this paper, we discuss the importance of using appropriate \textit{stopping criteria} and analyse the behaviour of a novel criterion based on the evolution of \textit{optimality criteria} in active graph-SLAM.
\end{abstract}


\end{frontmatter}

\section{Introduction} \label{S:1}

Actively exploring unknown environments has attracted the attention of the robotics community since the very first works appeared~\citep{bajcsy88, whaite97}. Basically, it encompasses controlling the movement of a robot which is capable of building a representation of the environment and localising itself on it. The latter are known as Simultaneous Localisation and Mapping (SLAM) and is the foundation of robotic exploration. See~\citet{thrun05},~\cite{durrant06},~\cite{cadena16} and references there in. Specifically, this article is concerned with graph-based SLAM, in which the problem is formulated using a graphical representation that encodes the robot poses in the graph nodes and the constraints between them (e.g. sensor measurements) in edges~\citep{grisetti10}.

Active SLAM (also referred to as robotic exploration) augments the above problem by also including a planning step. Thus, the autonomous agents must be capable of deciding where they should move next in order to improve the model of the environment (i.e. the map built and the localisation accuracy). Typically, this decision making problem has been viewed from an uncertainty-reduction standpoint, and divided in three phases for tractability: identification of possible actions, evaluation of their usefulness and execution of those considered as optimal. Solving this paradigm in finite time usually requires making several assumptions (e.g. maximum likelihood observations~\citep{platt10,valencia18,placed21} or a finite action set~\citep{carrillo18, placed20} but, even so, the second stage is computationally time-consuming: comparing the utility of two (sets of) actions involves reasoning over the probability density functions of the two random variables of interest, i.e. over the (usually high-dimensional) covariance or Fisher information matrices (FIM) of the beliefs.

In this sense, identifying the moment in which the exploration strategy is no longer adding information to the system is of utmost importance, in order to avoid unnecessary computational load. At that point, it could be changed or the exploration could simply be terminated. Besides obtaining irrelevant information, the continuous acquisition of redundant data has demonstrated to be detrimental for SLAM and may lead to map inconsistencies and even unrecoverable states. On the other hand, the ability to determine task completion is key for robots in real environments to be autonomous. Constraints typical of simulation (e.g. percent of coverage) are no longer valid in real-world applications in which the environment to explore is unknown by definition. The set of rules for deciding whether or not to terminate exploration is known as \textit{stopping criteria} (SC) or \textit{termination criteria}, and it was already identified by \citet{cadena16} as an open challenge more than five years ago. Despite the great impact of this work, however, no further research has been done in this field.

In this paper, we stress the importance of using task-driven SC towards autonomous robotic exploration, and thoroughly discuss why those commonly used in the literature are neither meaningful nor practical for that purpose. In addition, following the idea of using evolution of \textit{optimality criteria} as SC \citep{cadena16} and their fast computation via graph connectivity indices~\citep{placed21}, we present and analyse the performance of a novel criterion based on $D$-optimality. We seek to stimulate this field of research and encourage the use of meaningful criteria in future active SLAM works.

The rest of the manuscript is organised as follows. Section~\ref{S:2} presents the problem tackled. Section~\ref{S:3} contains a discussion on the different \textit{stopping criteria} used in the literature and their limitations. In section~\ref{S:4}, we describe and validate the proposed approach. Finally, the paper is concluded in Section~\ref{S:5}, also outlining future work.

\section{Background}\label{S:2}

\subsection{Active graph-SLAM} \label{SS:2a}

Active SLAM refers to the problem of controlling the movement of a robot which is performing SLAM in an unknown environment with the main objective of improving the quality of the SLAM estimates ($\mathcal{S}$): the map representation and the robot state. This paradigm requires agents to be capable of autonomously selecting the optimal actions to execute. Under the wider framework of Partially Observable Markov Decision Processes (POMDP), it can be formulated as a decision making problem in which a decision, $\delta$, is defined by the mapping:
\begin{equation}
    \delta : \mathcal{P}(\mathcal{S})\to\mathcal{A} \label{eq:1}
\end{equation}
\noindent That is, it maps elements from the space of probabilities over the robot and map states (i.e. beliefs, $\mathcal{P}(\mathcal{S})$) to the action space, $\mathcal{A}$.

In general, a decision will be preferred over another if it is more useful. Utility functions ($\mathcal{U}$) measure this usefulness by mapping the complex spaces $\mathcal{P}(\mathcal{S})$ and $\mathcal{A}$ to the real scalar space, in which comparison can be made directly:
\begin{equation}
    \mathcal{U} : \mathcal{P}(\mathcal{S})\times\mathcal{A}\to \mathbb{R} \label{eq:2}
\end{equation}

Thus, decision making in active SLAM can be reduced to the following optimisation problem:
\begin{equation}
    \textbf{a}^* = \arg\max_{\textbf{a}\in\mathcal{A}} \ \mathcal{U} \label{eq:3}
\end{equation}
\noindent where $\textbf{a}^*$ is the optimal set of actions to be executed.

Traditionally, active SLAM has been divided in three differentiated phases for the ease of its resolution~\citep{makarenko02}, which are directly linked to eq.~(\ref{eq:1})-(\ref{eq:3}). During first step, and just for computational purposes, the (usually finite) set of possible destinations the robot could travel to is identified (e.g. via frontier detection). Second stage deals with utility computation. The effect of executing the candidate actions is evaluated by quantifying the expected uncertainty in the two random variables. Information-theoretic (IT) approaches are the most common and are usually based on the concept of entropy~\citep{stachniss03,valencia12,palomeras19}. On the other hand, methods based on Theory of Optimal Experimental Design (TOED) directly compute utility from the variance of the variables of interest. \textit{Optimality criteria} quantify utility by mapping elements from the space of real symmetric matrices (covariance or FIM) to $\mathbb{R}$. \textit{D}-optimality has been considered the most fruitful criterion~\citep{rodriguez18}, and its modern formulation is as follows~\citep{pukelsheim06}:
\begin{align}
    D{\text -} opt &\triangleq \exp \left(\frac{1}{d} \sum_{k=1}^d \log(\lambda_k) \right) \label{eq:dopt}
\end{align}
\noindent where $\Lambda=(\lambda_1,\dots,\lambda_d)$ are the eigenvalues of the matrix under study, with dimension $d$. Finally, during the third step, the frontier with highest utility is selected as optimal and the set of actions to reach it is executed. The entire sequence would be repeated until a certain SC was met.

This paper is particularly focused on active graph-SLAM methods, in which SLAM is formulated using a graph representation where nodes represent the robot poses, and edges encode the constraints between them, usually in the form of a FIM. For the $j$-th edge, it will be $\boldsymbol{\Phi}_j\in\mathbb{R}^{\ell\times\ell}$ with $\ell$ the dimension of the state vector (e.g. 3 for 2D SLAM). These methods rely on the insight that the map representation (usually in the form of an occupancy grid) can be retrieved once the robot states have been properly estimated~\citep{montemerlo03, grisetti10}. Evaluation of \textit{optimality criteria} in such cases involves computing the eigenvalues of the system's FIM, $\textbf{Y}\in\mathbb{R}^{n\ell\times n\ell}$, with $n$ the number of nodes in the graph. It should be noted that the matrix dimensions grow as the trajectory does, and that the map information must somehow be included in $\textbf{Y}$ to effectively address the task under study. See~\cite{placed21} for a more detailed discussion on active graph-SLAM and $\textbf{Y}$.

\subsection{Exploiting the Graphical Structure in Active SLAM} \label{SS:2b}

Evaluation of utility is costly and quickly becomes intractable for online approaches. It usually represents the main bottleneck in active SLAM, especially if TOED-based. Thus, most works resort to IT, sparse information matrices~\citep{indelman15} or sparsified representations~\citep{carrillo18, elimelech19}.

Regarding computational complexity, the prominent works by~\cite{khosoussi14, khosoussi19} show that classical \textit{optimality criteria} are closely linked to the connectivity of the underlying pose-graph. Thus, instead of evaluating the whole FIM, decision making in active graph-SLAM can be done in a fraction of the time\textemdash up to an order of magnitude faster~\citep{kitanov19,placed21}\textemdash by studying the graph connectivity indices.~\cite{placed21b,placed21} formulate the existing relationship between modern \textit{optimality criteria} of the FIM and those of the weighted Laplacian, for the general case of active graph-SLAM. Specifically, for $D$-optimality:
\begin{align}
    D{\text -} opt(\textbf{Y}) &\approx \left( n\ t(\boldsymbol{\mathcal{G}}_\gamma)\right)^{\frac{1}{n}}\label{eq:dopt_graph}
\end{align}
\noindent where $\boldsymbol{\mathcal{G}}_\gamma$ denotes a pose-graph in which each edge is weighted with the same criterion to be estimated, i.e. $\gamma_j=D\text{-}opt\left(\boldsymbol{\Phi}_j\right)$, and $t(\boldsymbol{\mathcal{G}}_\gamma)$ is the weighted number of spanning trees of the graph, equal to any cofactor of its weighted Laplacian. See~\cite{placed21}.

\section{Limitations of the Existing Stopping Criteria} \label{S:3}

\renewcommand\tabularxcolumn[1]{m{#1}}
\renewcommand{\arraystretch}{1.3}
\begin{table*}[t!]
    \begin{center}
        \caption{Example works, formulation basis and limitations of the most relevant SC in literature.}
        \label{tab:sc_literature}
        \begin{tabularx}{0.99\textwidth}{m{\dimexpr 0.15\textwidth-2\tabcolsep}|m{\dimexpr 0.09\textwidth-2\tabcolsep}||m{\dimexpr 0.33\textwidth-3\tabcolsep}||m{\dimexpr 0.12\textwidth-2\tabcolsep}|>{\raggedright\arraybackslash}m{\dimexpr 0.30\textwidth-2\tabcolsep}}
            \multicolumn{2}{l||}{\textbf{Stopping Criterion}} & \textbf{Works} &\textbf{Formulation} & \textbf{Limitations} \\ \hline \hline
            \multicolumn{2}{l||}{Temporal} & \cite{leung08},~\cite{valencia12},~\cite{carrillo18}~and~\cite{placed21} &Time space &\textbullet~No~task~completion \textbullet~Must~be~set~experimentally (scenario- and robot-dependent) \textbullet~Not~comparable~across~systems\\ \hline
            \multirow{6}{*}{Spatial} & Frontiers &\cite{yamauchi99},~\cite{korb18}~and~\cite{pimentel18} & &\multirow{6}{\dimexpr 0.30\textwidth-2\tabcolsep}{\textbullet~No~task~completion (coverage only) \textbullet~Difficult~to~compare~across~systems \textbullet~Unobservable~and unreachable~areas \textbullet~Most~times~require~prior~knowledge of~the~environment~or~human interaction}\\ \cline{2-3}
                & Coverage &\cite{pham13},~\cite{amigoni13},~\cite{lenac16},~\cite{palomeras19},~\cite{chen20} and~\cite{xu22} &State space & \\ \cline{2-3}
                & Other & \cite{bircher18}~and~\cite{suresh20} & &\\ \hline
            \multirow{2}{*}{Uncertainty-based} & IT  &\cite{simmons00},~\cite{stachniss04},~\cite{salan14},~\cite{chen19},~\cite{ghaffari19},~\cite{gomez19}~and~\cite{deng20}&Information space &\textbullet~No~task~completion (coverage only) \textbullet~Difficult~to~compare~across~systems \textbullet~Unobservable~and unreachable~areas \textbullet~May~require~prior~knowledge~of~the environment \\ \cline{2-5}
               & TOED & --- &Task space & \textbullet~Computational and formulation complexity
        \end{tabularx}
    \end{center}
\end{table*}

Most active SLAM works in the literature resort to spatial or temporal constraints as SC. The former can be used only because they use known (simulation) environments for which the complete size of the map is available, and the latter typically encode physical constraints of the robot (e.g. battery) rather than those of the variables of interest. Both of them must be experimentally designed for a specific scenario and can be used neither in unknown environments, nor across systems. Table~\ref{tab:sc_literature} illustrates the preferred choice of SC in some relevant works. It is easy to notice how temporal and spatial (i.e. geometric) criteria monopolise the attention in the literature.

Limiting the amount of time for an exploration task has been used extensively~\citep{leung08,valencia12,carrillo18,placed21} for two main reasons: its simplicity and because comparing the performance of diverse agents in the same time horizon is straightforward. This criterion does not directly require previous knowledge about the environment, although for exploration results to be relevant, they have to be accordant. In any case, it (i) does not guarantee exploration of the entire environment or low uncertainty estimates (i.e. task-completion), (ii) is dependent on the scenario and the robot configuration, and (iii) often needs to be set experimentally (e.g. by manually exploring the environment before testing the approach).

The simplest geometric SC broadly used in the literature is the non-existence of locations to explore, which usually translates in the absence of (clusters of) frontiers~\citep{yamauchi99, korb18,pimentel18}. A second relevant geometric constraint is the desired coverage (e.g. $90\%$). For its use, by definition, it is mandatory to know the size or even the full map of the environment. Despite being in conflict with active SLAM definition, it is often used \citep{pham13,amigoni13,lenac16,palomeras19,chen20}. A recent example can be found in~\citep{xu22}, where they propose a unified framework to benchmark robotic exploration, although assume known environments and base efficiency on coverage. Other geometric-related SC rely on limiting the size of the exploration trees~\citep{bircher18}, or on qualitative metrics that require human supervision~\citep{suresh20}. In contrast to temporal SC, all of the above allow checking the completion of the task of covering a surface or volume, since they are formulated over the state space. Nevertheless, two insights are worth noting: (i) the fundamental aspect in active SLAM of lack of prior information about the environment is violated, and (ii) they do not assess the active SLAM task but a coverage one, so they have no concern about the quality of the estimates. In fact, they do not even monitor the robot localisation.

Following Information Theory (IT) and the well-known (approximation) relationship between the entropy of an occupancy map and the number of unknown cells on it~\citep{stachniss09}, a number of works~\citep{simmons00,chen19,deng20} soon reformulated the aforementioned spatial SC in the so-called information space (or belief space). That is, the place where the probability density functions of the random variables naturally live. See~\cite{barraquand95} and~\cite{prentice09}. These probability distributions reflect knowledge about the system, and related metrics typically measure the entropy reduction in the posterior (map) distribution. ~\cite{gomez19} offer a fresher look at spatial criteria, using the concept of ``interesting" frontiers, that is, frontiers with an expected utility above a defined threshold (e.g. expected Information Gain, IG). Determining this threshold must be done experimentally, though, and is scenario-specific.~\cite{ghaffari19} use a saturation information value over the current (and not necessarily complete) map, which reflects the confidence in the representation.~\cite{stachniss04} observe that considering only a constant upper bound on the map information can lead to repeatedly acquiring the same information, and that is indeed detrimental to the SLAM performance.~\cite{salan14} also realise that coverage SC are impractical, even when formulated over the information space: some cells can be unobservable or even unreachable. Instead, they propose to conduct exploration until there are no possible configurations that maximised information.

It is essential to note that, despite fast, information-driven metrics do not allow to check for task-completion in any case, since they are not related to the task of active SLAM but to that of coverage. Consider the example case in which all potential locations to travel to would result in low informative returns (e.g, entropy or IG below a predefined threshold or even zero). According to information-driven SC, exploration would be finished, but not necessarily implying that the two variables of interest were perfectly nor even well estimated, since the information units are unrelated to the physical meaning of the task (e.g. $\SI{}{\meter\squared}$ for linear variance).
\section{Towards Meaningful Stopping Criteria}\label{S:4}

The use of metrics (either or raw or their evolution) that stem from TOED as \textit{stopping criteria} has been identified as promising many times \citep{cadena16, lluvia21}, although no research has been done in the topic. In contrast to IT metrics, they are task-driven, that is, they evaluate uncertainty in the task space (i.e. variance of estimates) and therefore allow to directly check if a given set of actions improves the task or if the estimations are sufficiently accurate. This also implies that they can be compared across systems. Their main drawback is the high computational load that evaluating large covariance matrices has traditionally required. However, thanks to the fast computation of graph connectivity indices and their equivalence to \textit{optimality criteria} in active graph-SLAM (see Section \ref{SS:2b}), we propose their evolution along exploration as SC.

A well-conditioned SC has to account for both variables of interest in active SLAM: the robot localisation and map representation. Revisiting Section~\ref{SS:2a}, graph-based SLAM builds upon the idea that the map can be straightforwardly retrieved once the robot is properly localised. Therefore, we propose the following SC based on the evolution of the robot's uncertainty (via graph's $D$-optimality) and the mapped area:
\begin{equation}
    \Gamma = \Delta\mathcal{U}+|\Delta A| < \Gamma_{th}
\end{equation}
\noindent where $\mathcal{U}$ is the utility function (see equation \ref{eq:dopt_graph}), $A$ the known area in the map and $\Delta$ denote percentage variations w.r.t. previous active SLAM steps. The area has been restricted to absolute variations, as otherwise, a loop closure could trigger the criterion after updating the map. Note that the map's accuracy is embedded in $\Delta\mathcal{U}$ given the graph-SLAM basis.

Should $\Gamma$ drop below a certain threshold ($\Gamma_{th}$) during a given action window, $w$, active SLAM would be terminated as no information would be added to the system. Thus, for a fixed map size, if the robot keeps gathering the same information over and over (i.e. $\Delta\mathcal{U}\lesssim 0$), it will stop. Otherwise, exploration will continue even when information decreases (pure exploration) until the information loss exceeds the gain in the map area. This would indicate that a different exploration strategy is required, rather than that exploration has finished. Therefore, the proposed SC depends on two parameters: the minimum increment $\Gamma_{th}$, and the window in which evaluations are performed. The former is related to task-completion, and the smaller it is, the most complete, accurate and time-expensive the task will be. The latter serves to check whether it is a stationary or transient regime: short horizons (e.g. 1 action) should not be used to check for termination.

\begin{figure}
    \centering
    \includegraphics[max height=0.45\linewidth,max width=\linewidth]{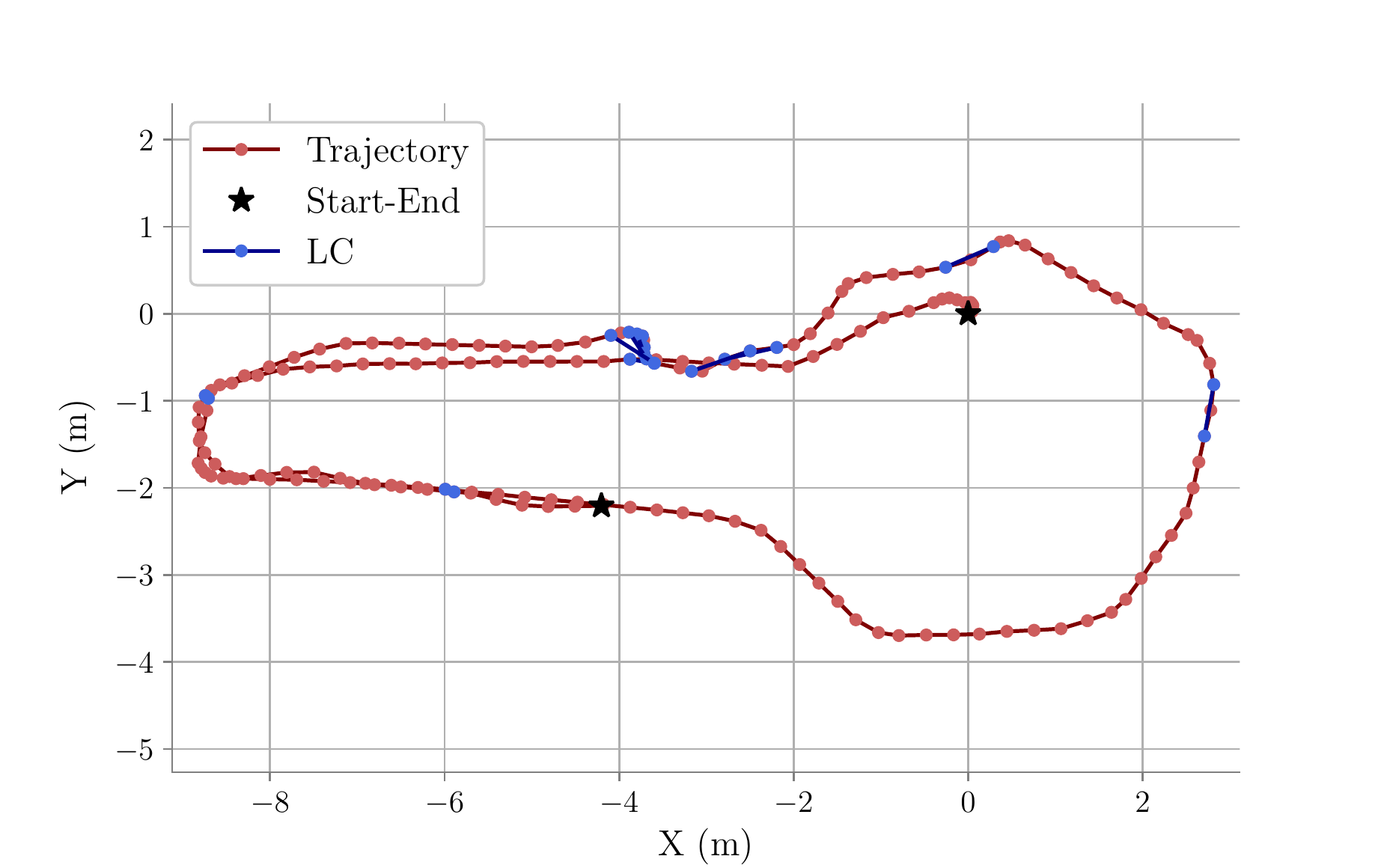}
    \caption{Trajectory of the active SLAM example experiment.}
    \label{fig:sc_traj}
\end{figure}

\begin{figure}
    \centering
    \includegraphics[max height=0.5\linewidth,max width=\linewidth]{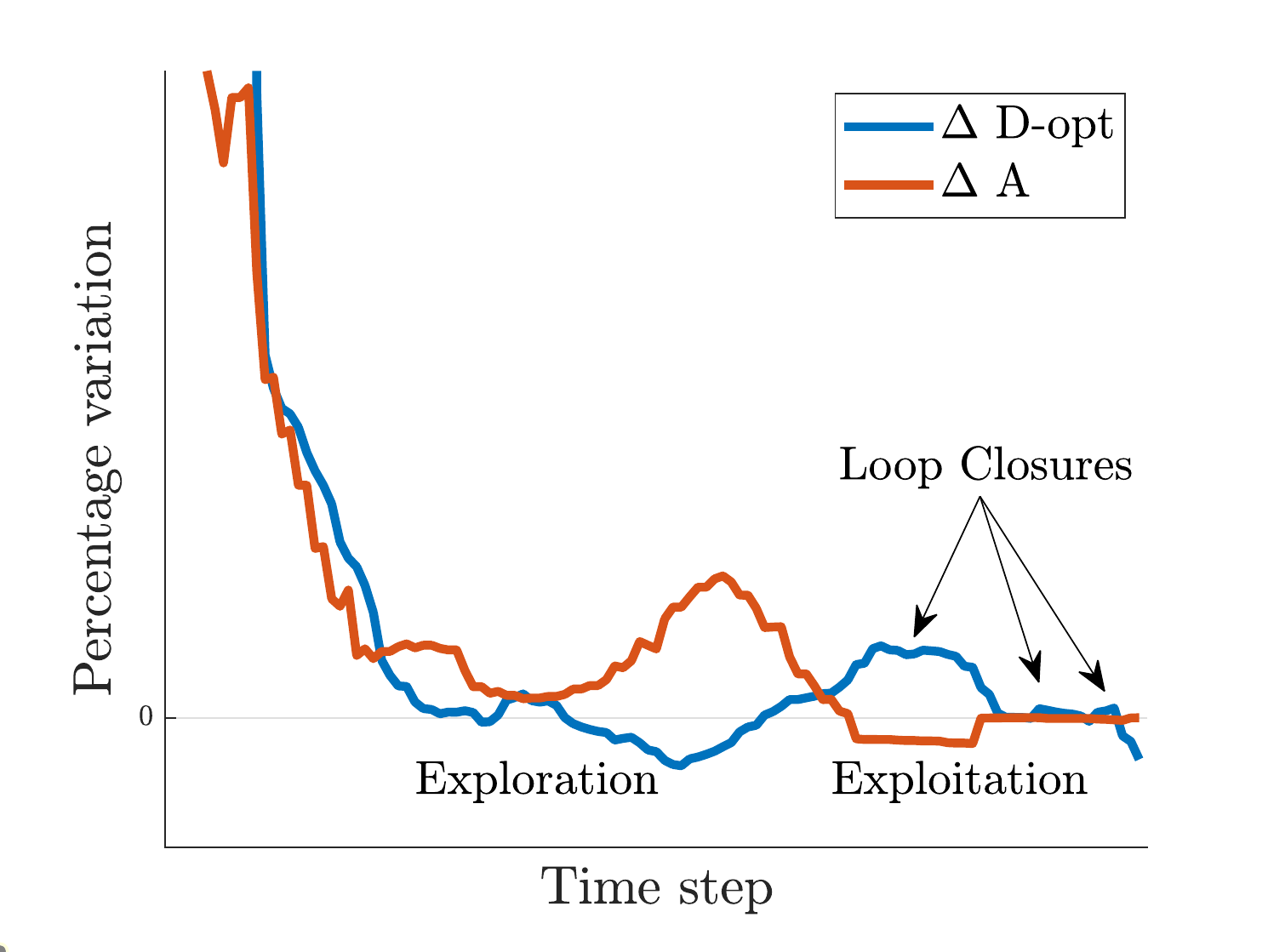}
    \caption{Evolution of $\Delta D$-$opt$ and $\Delta A$ during the active SLAM example experiment.}
    \label{fig:sc_variation}
\end{figure}

Let us consider a simple example of active SLAM to illustrate the typical trend of both terms. Figure \ref{fig:sc_traj} contains its trajectory (red dots correspond to graph nodes and red lines to odometry edges) and loop closures (blue). Figure \ref{fig:sc_variation} shows the percentage variation in $D$-optimality (blue) and area (red) during the task. Note that $D\text{-}opt$ is computed over $\textbf{Y}$ and thus denotes knowledge over the robot's localisation. Two stages can be clearly distinguished, also labelled in the figure: exploration and exploitation. During the first stage, there is a large initial increase in the known area and localisation information. As exploration continues, new areas decrease and the robot's uncertainty gradually increases. At a certain point, the robot starts to revisit known places, entering the second stage. During exploitation, the area no longer increases, and may even decrease after closing a loop due to map updates. On the other hand, localisation uncertainty continues to decrease as loops are closed. However, once the first few loop closures occur, there is no further improvement. Both increments reach a constant regime near zero and after this, if the robot keeps gathering the same data from the environment, there is a loss of information (see final part of Figure \ref{fig:sc_variation}). This regime represents the point at which exploration should be terminated or the utility function changed.

\begin{table*}[t!]
    \centering
    \caption{Results of exploration in AWS scenarios with agents with different SC. They have been ordered by the time consumed. $\dagger$ denotes values are explicitly fixed by the criterion.}
    \begin{tabular}{c||c||c||c|c|c||c|c|c|c}
        \textbf{Scenario} & \textbf{SC} & \textbf{Time} (\SI{}{\second}) & \textbf{Area} (\SI{}{\meter\squared}) & \textbf{Coverage} ($\%$) & \textbf{mRMSE} (\SI{}{\meter}) & \textbf{n} & \textbf{d} & \textbf{Opt.} & $\textbf{D\text{-}opt}$\\ \hline
        \multirow{4}{*}{Bookstore} & Temporal (\SI{10}{\min}) & $\dagger600$ & 155.04 & 86.01 & 0.22 & 492 & 2.48 & 3 & 3372.58 \\ \cline{2-10}
        & \textit{Ours} & 1018 & 159.56 & 88.52 & 0.11 & 867 & 2.74 & 6 & 3862.94 \\ \cline{2-10}
        & Coverage ($90\%$) & $\infty$ & -- & $\dagger90$ & -- & -- & -- & -- & -- \\ \cline{2-10}
        & Coverage ($99\%$) & $\infty$ & -- & $\dagger99$ & -- & -- & -- & -- & -- \\ \hline \hline
        \multirow{4}{*}{House} & Coverage ($90\%$) & 301 & 149.53 & $\dagger90$ & 0.72 & 236 & 2.49 & 0 & 2869.24 \\ \cline{2-10}
        & \textit{Ours} & 482 & 155.40 & 93.53 & 0.23 & 331 & 2.45 & 1 & 2941.25 \\ \cline{2-10}
        & Temporal (\SI{10}{\min}) & $\dagger600$ & 155.48 & 93.58 & 0.24 & 382 & 2.40 & 2 & 2909.94 \\ \cline{2-10}
        & Coverage ($99\%$) & $\infty$ & -- & $\dagger99$ & -- & -- & -- & -- & --
    \end{tabular}
    \label{tab:stopping_criteria}
\end{table*}

We are aware that evaluating variations on $D$-$opt$ rather than raw values does not allow to directly constrain the robot's variance. Still, contrary to usual SC in the literature, it does neither imply prior knowledge of the environment nor human interaction during the task, takes into account both variables of interest and is transferable across systems. 

\subsection{Experimental Validation}\label{S:4a}

\begin{figure*}[t!]
    \centering
    \subfloat[AWS bookstore environment and the robot in Gazebo.]{%
        \centering
        \includegraphics[max height=3.8cm,max width=\linewidth]{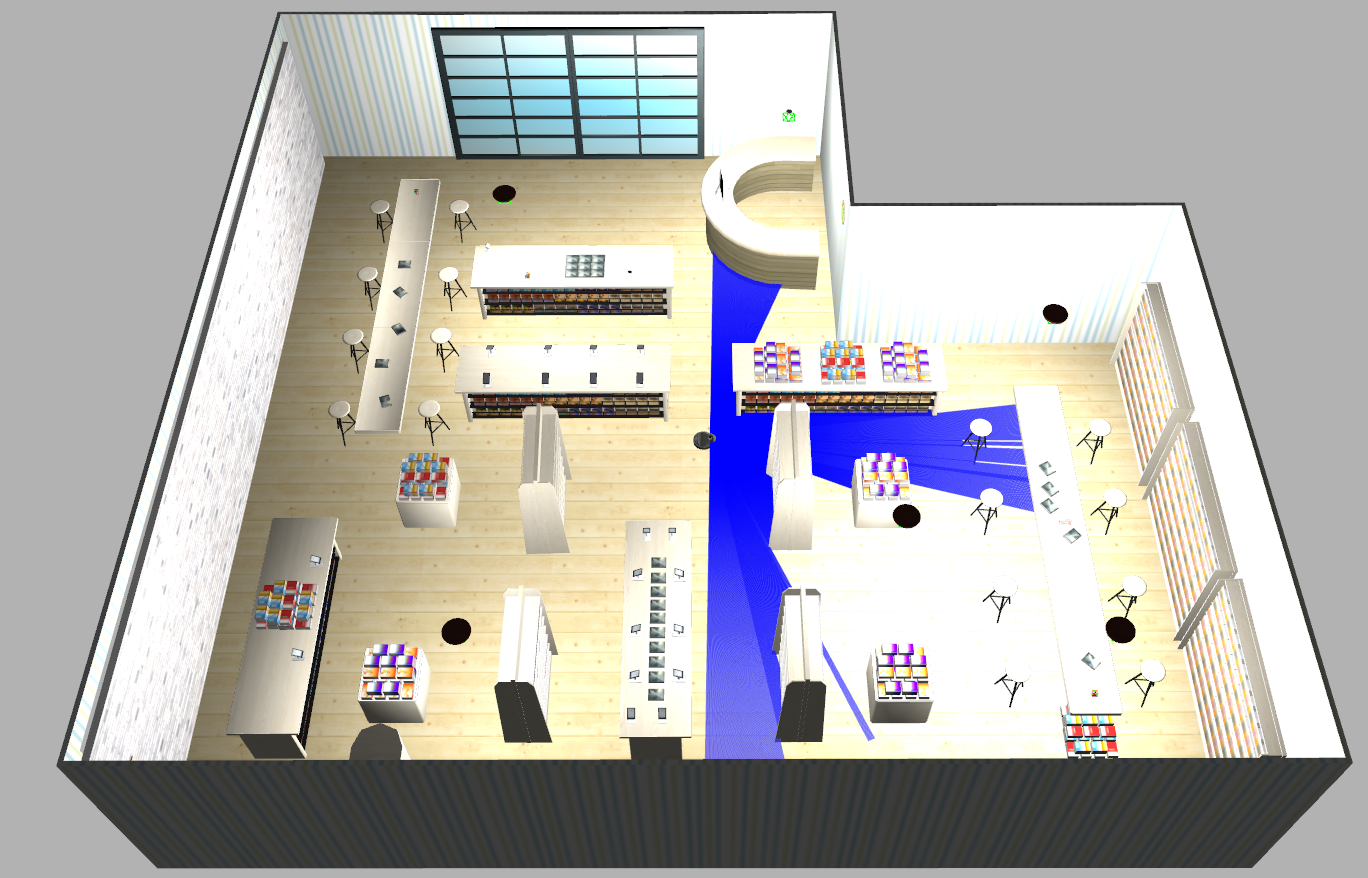}
        \label{fig:aws_bs}
    } \qquad
    \subfloat[Temporal criterion.]{%
        \centering
        \includegraphics[max height=3.8cm,max width=\linewidth]{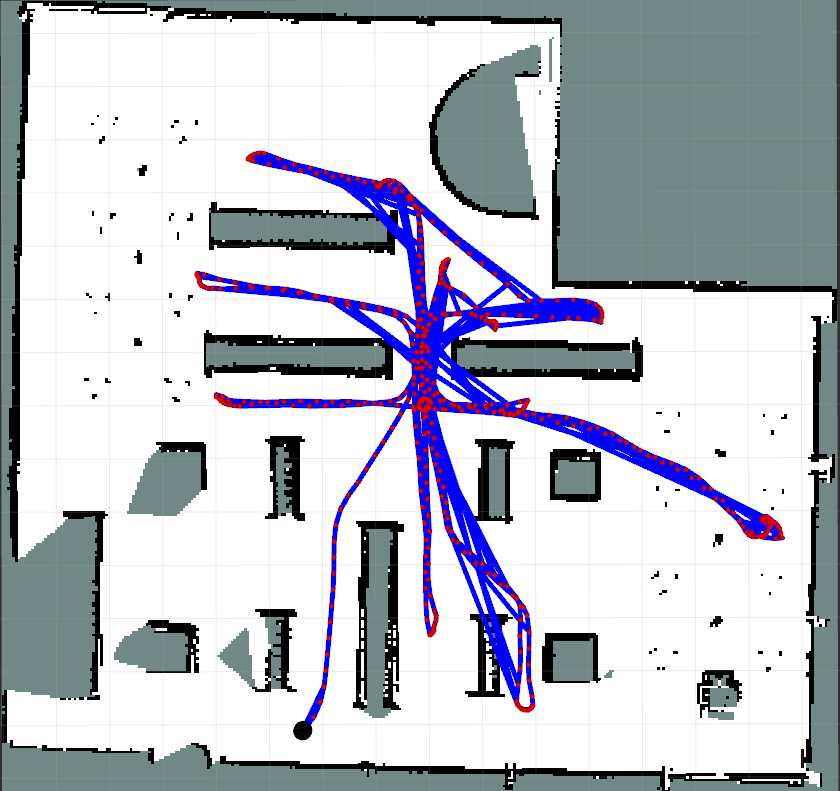}
        \label{fig:map_bs2}
    } \qquad
    \subfloat[Task-driven criterion (\textit{ours}).]{%
            \centering
            \includegraphics[max height=3.8cm,max width=\linewidth]{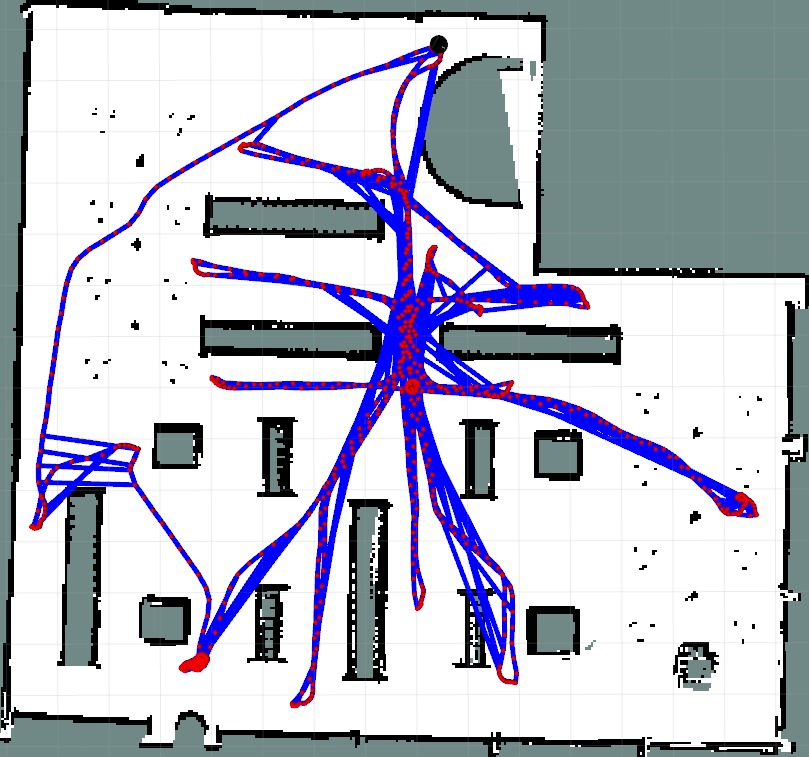}
            \label{fig:map_bs1}
    } \\
    \subfloat[AWS house environment and the robot in Gazebo.]{%
            \centering
            \includegraphics[max height=6cm]{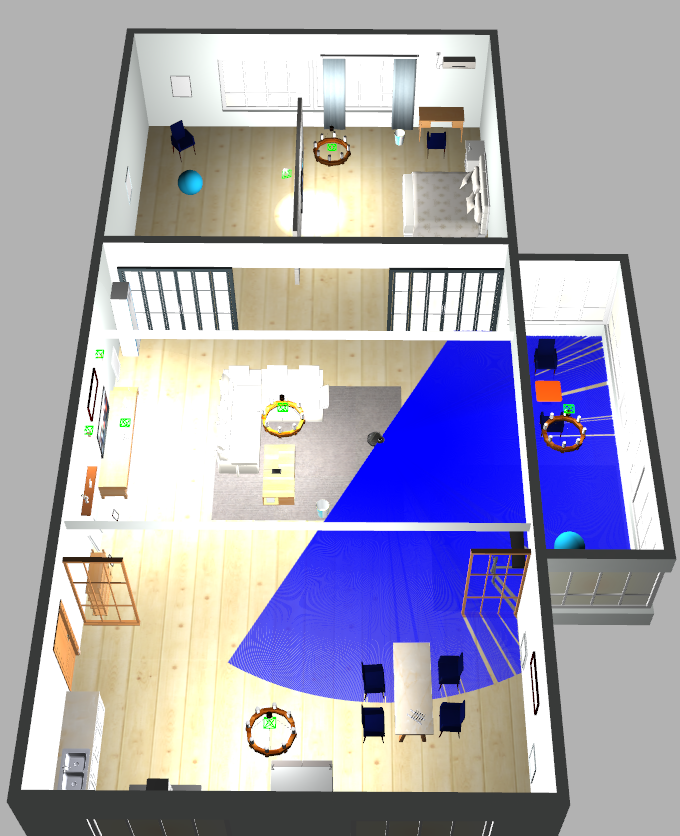}
            \label{fig:aws_house}
    } \quad
    \subfloat[Coverage criterion ($90\%$).]{%
        \centering
        \includegraphics[max height=6cm,max width=3.6cm]{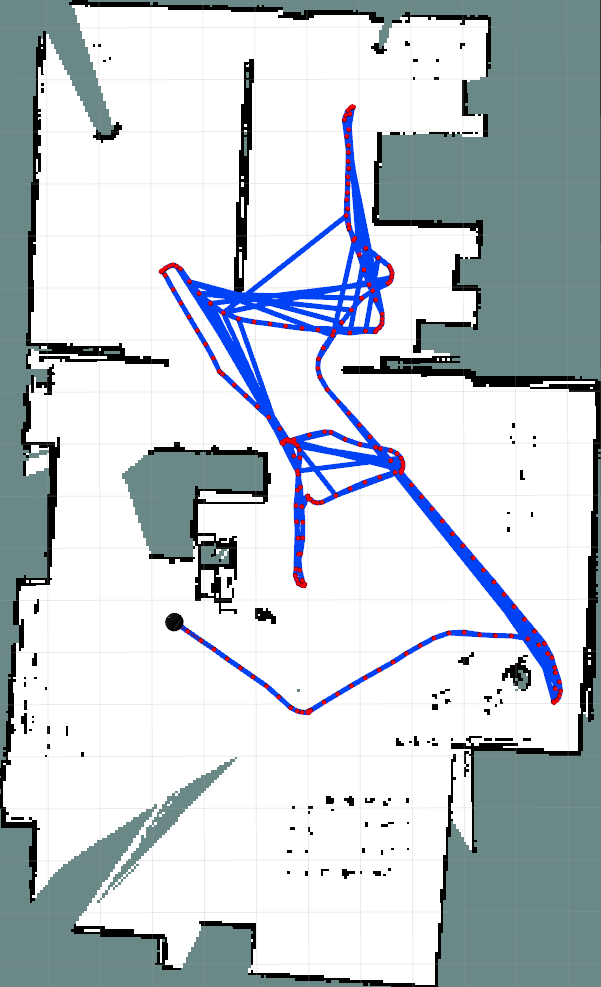}
        \label{fig:map_house2}
    } \quad
    \subfloat[Task-driven criterion (\textit{ours}).]{%
        \centering
        \includegraphics[max height=6cm,max width=3.6cm]{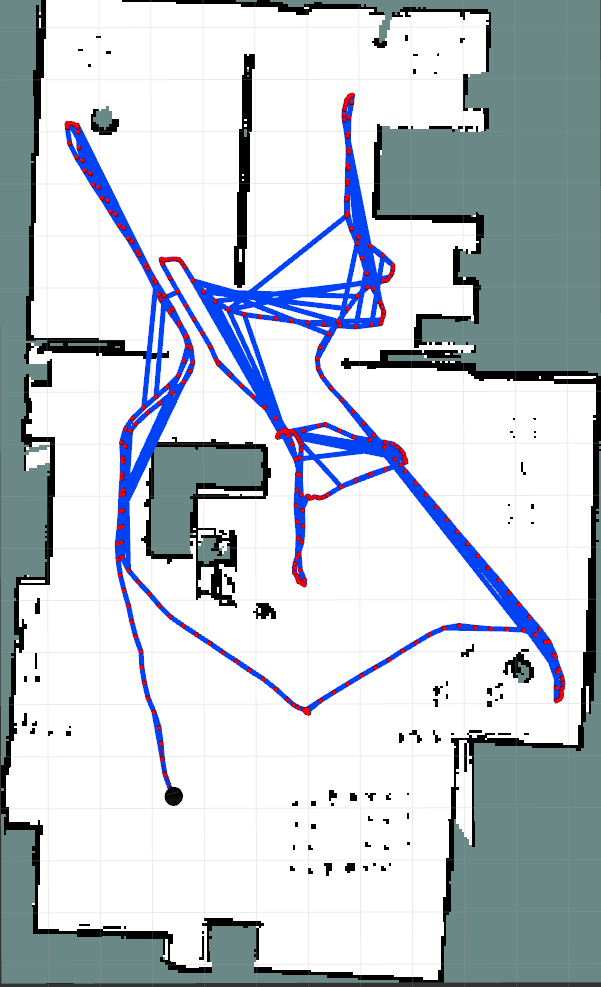}
        \label{fig:map_house3}
    } \quad
    \subfloat[Temporal criterion.]{%
            \centering
            \includegraphics[max height=6cm,max width=3.6cm,]{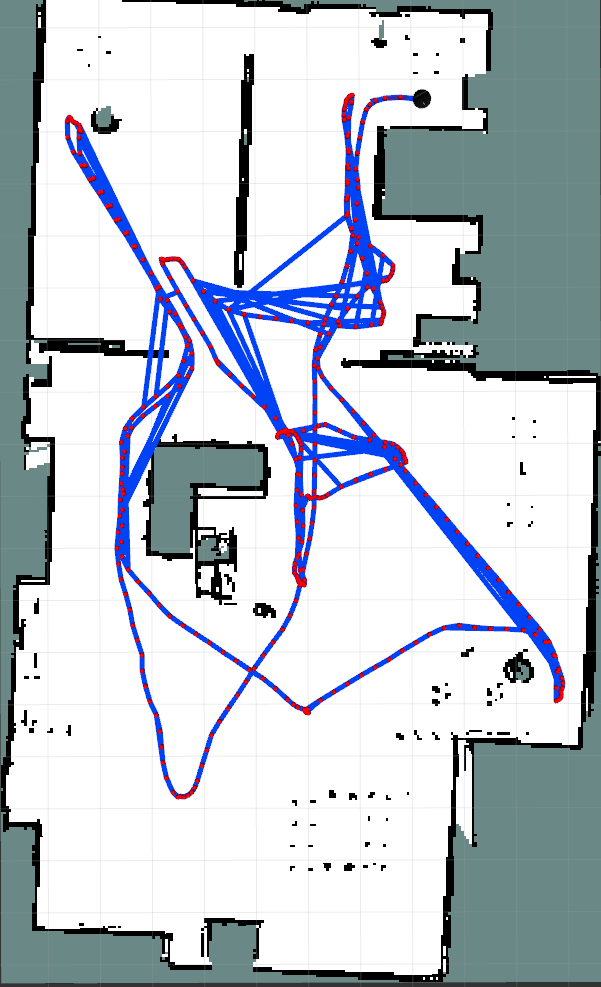}
            \label{fig:map_house1}
    }
    \caption{View of the AWS bookstore (a) and house (d) environments in Gazebo. Also, the maps and pose-graphs built in one trial using temporal, coverage and TOED-based \textit{stopping criteria} in both environments.Results for coverage SC have been omitted when unfeasible. }
    \label{fig:results_sc}
\end{figure*}

In order to test the proposed criterion, active SLAM was performed in two scenarios in Gazebo, which are slightly modified versions of AWS bookstore and house worlds (e.g., they are now closed environments). See figures~\ref{fig:aws_bs} and~\ref{fig:aws_house}. The robot is a wheeled platform equipped with a laser sensor with \SI{180}{\degree} field of view, \SI{5}{\meter} range and $1500$ beams in each scan. The maximum linear and angular velocities of the robot are \SI{0.2}{\meter\per\second} and \SI{0.8}{\radian\per\second}, respectively. Active SLAM framework is the same as in~\citep{placed21}, which is publicly available.

The proposed experiment consists of determining and comparing the moment at which different SC decide to stop exploration. To make a fair comparison, different criteria are evaluated during the same exploration run. That is, as the autonomous agent explores the environment, SC are checked and if any of them is met, results for that criterion are extracted. Then, the exploration is resumed. Apart from the task-driven criterion ($\Gamma_{th}=2\%$, $w=3$), we used a time-based one (\SI{10}{\min}) and two coverage conditions: $90$ and $99\%$, following the somewhat standardised values~\citep{xu22}.  Table~\ref{tab:stopping_criteria} shows results (mean over 2 trials) of all agents in both scenarios. It contains the time elapsed until the stopping conditions were met, map metrics (area explored, coverage, max. RMSE) and graph metrics (size, average node degree, number of graph optimisations and final graph \textit{D-opt}). Note that coverage conditions assume the environment known, and that the time has been arbitrarily selected. Fig.~\ref{fig:results_sc} shows the maps and pose-graphs at the moment of fulfilment of the different criteria.

In the bookstore environment, time-based criterion is the first to be met: after 10 minutes of exploration, the agent knows $86\%$ of the scenario and maximum error of the mapped area is \SI{22}{\centi\meter}. Almost 7 minutes later, the proposed criterion is satisfied. Despite the similar coverage (only $2.5\%$ higher), the number of graph optimisations has doubled and map error has halved. Also, the robot localisation information has increased by $15\%$. Coverage conditions are never satisfied, as there are many unobservable or unreachable areas for the robot.

On the other hand, the $90\%$ coverage criterion is the first to be fulfilled in the second scenario. However, at that time the accuracy of the robot localisation and of the built map are extremely low. In addition, many interesting areas have not yet been explored (see Figure~\ref{fig:results_sc}). After 8 minutes of exploration, the TOED-based criterion is met: not only $3.5\%$ more of the environment has been explored, but also known areas have been exploited: map maximum error has been reduced threefold and the localisation information has increased. Finally, after 10 minutes, the temporal criterion triggers, which, regarding the map and coverage, has very similar results to our criterion. This experiment illustrates how repeatedly acquiring the same data (over-exploitation) can be indeed detrimental for the SLAM algorithm: despite a higher number of loop closures, the robot localisation and the map are less accurate and the graph is not as well connected (see last columns of Table~\ref{tab:stopping_criteria}). Again, the $99\%$ coverage criterion is never met.

The proposed task-driven criterion achieves, in all cases, a relevant behaviour for active SLAM, outperforming other studied criteria. On the one hand, coverage criteria require prior knowledge of the environment and, even then, may fail. On the other hand, time-based criteria either require prior knowledge or are impractical: large times may lead to over-exploitation while short may fail to visit areas of interest. Our criterion succeeds in autonomously identifying the moment in which exploration is no longer adding relevant information, without requiring prior knowledge or manual tuning beyond the specification of the two parameters it depends on.

\section{Conclusions}\label{S:5}

In this manuscript, we have studied the often overlooked problem of autonomously deciding when robotic exploration should terminate. First, we have thoroughly reviewed the use of existing \textit{stopping criteria} in the literature, highlighting the assumptions on which they are based and their limitations for performing active SLAM in real environments. Motivated by this, and following the recent work on fast computation of \textit{optimality criteria} in active graph-SLAM, we have proposed a TOED-based \textit{stopping criterion}, which autonomously decides when exploration is not adding further information to the SLAM system. Experimental results show the relevance of this criterion for active SLAM and how it outperforms others used in the literature. This seminal work aims to stimulate the use of task-driven \textit{stopping criteria} in active SLAM, rather than coverage or temporal metrics.

Future work will aim to study the use of raw \textit{optimality criteria} rather than their evolution, and will evaluate their use to decide to change the exploration strategy (i.e. switch the utility function) instead of just terminating exploration.

\bibliography{references}

\end{document}